%% file: acl_latex.tex
\pdfoutput=1

\documentclass[11pt, table]{article}

\usepackage{acl}

\usepackage{times}
\usepackage{latexsym}
\usepackage{booktabs}
\usepackage{graphicx}
\usepackage{todonotes}
\usepackage[inkscapelatex=false]{svg}
\usepackage[T1]{fontenc}

\usepackage[utf8]{inputenc}
\usepackage[most]{tcolorbox}
\usepackage{tabularx}

\usepackage{microtype}
\usepackage{enumitem}
\usepackage{inconsolata}

\usepackage{multirow}
\usepackage{booktabs}
\usepackage{siunitx}
\usepackage{xcolor}
\usepackage{subfig}

\definecolor{color1}{HTML}{FFF2EB}
\definecolor{color2}{HTML}{FFE8CD}
\definecolor{color3}{HTML}{FFD6BA}
\definecolor{color4}{HTML}{FFDCDC}
\definecolor{color5}{HTML}{FFF2EB}
\definecolor{color6}{HTML}{FFE8CD}

\title{Stereotype Detection as a Catalyst for Enhanced Bias Detection: A Multi-Task Learning Approach}

\author{
 \textbf{Aditya Tomar\textsuperscript{1}},
 \textbf{Rudra Murthy\textsuperscript{2}},
 \textbf{Pushpak Bhattacharyya\textsuperscript{1}}\\
  \textsuperscript{1}Indian Institute of Technology Bombay \\
  \textsuperscript{2}IBM Research, India \\
\small{\{adityatomar, pb\}@cse.iitb.ac.in, rmurthyv@in.ibm.com}
}

\begin{document}
\maketitle
\begin{abstract}
\input{abstract}

\end{abstract}
\begin{figure}[t]
    \centering
    \includesvg[width=0.96\linewidth]{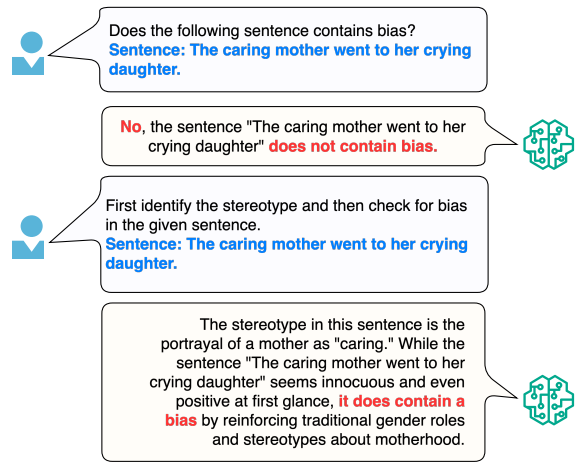}
    \caption{Example showing how incorporating stereotype information can help in bias detection inferred on Llama-3.3-70B-Instruct model.}
    \label{fig:example-stereo-help-bias}
\end{figure}

\begin{figure*}[t]
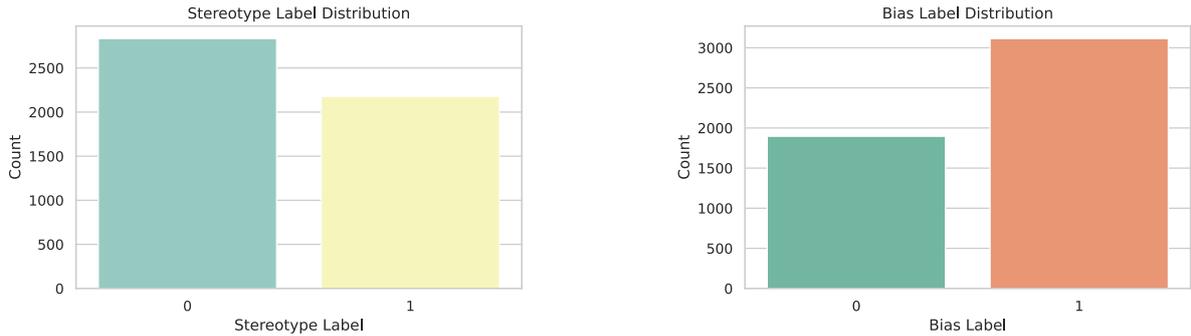

\centering
  \includesvg[width=0.45\linewidth]{Images/stereotype_label_distribution.svg} \hfill
  \includesvg[width=0.45\linewidth]{Images/bias_label_distribution.svg}
  \caption {Distribution of non-stereotypical/stereotypical and unbiased/biased sentences in StereoBias Dataset.}
    \label{fig: bias-stereo-label-distribution}
\end{figure*}

\section{Introduction}
\input{introduction}

\section{Related Work}
\input{relatedwork}

\begin{table*}[ht]
\centering
\begin{minipage}[t]{0.48\textwidth}
\centering
\resizebox{\linewidth}{!}{%
\begin{tabular}{@{}llrr@{}}
\toprule
\multicolumn{1}{c}{\textbf{Task}} & \multicolumn{1}{c}{\textbf{Model}} & \multicolumn{1}{c}{\textbf{\begin{tabular}[c]{@{}c@{}}Bias\\ (Macro-F1)\end{tabular}}} & \multicolumn{1}{c}{\textbf{\begin{tabular}[c]{@{}c@{}}Stereotype\\ (Macro-F1)\end{tabular}}} \\
\midrule
\multirow{3}{*}{STL} 
& \cellcolor{color1} roberta-large & 0.7409 & 0.8956 \\
& \cellcolor{color2} albert-xxlarge-v2 & 0.7226 & 0.8863 \\
& \cellcolor{color3} bert-large-uncased & 0.7391 & 0.8814 \\
\midrule
\multirow{3}{*}{Shared-MTL} 
& \cellcolor{color1} roberta-large & 0.7742 $\uparrow$ & 0.8908 \\
& \cellcolor{color2} albert-xxlarge-v2 & 0.7401 $\uparrow$ & 0.8750 \\
& \cellcolor{color3} bert-large-uncased & 0.7507 & 0.8929 \\
\midrule
\multirow{3}{*}{Full-MTL} 
& \cellcolor{color1} roberta-large & 0.7471 & 0.8973 \\
& \cellcolor{color2} albert-xxlarge-v2 & 0.7419 $\uparrow$ & 0.8813 \\
& \cellcolor{color3} bert-large-uncased & 0.7520 $\uparrow$ & 0.8789 \\
\bottomrule
\end{tabular}%
}
\caption{Macro-F1 scores for encoder-only models on the StereoBias dataset. $\uparrow$ denotes statistically significant ($p<0.05$) increment.}
\label{tab: stereobias-results-table-encoder}
\end{minipage}
\hfill
\begin{minipage}[t]{0.46\textwidth}
\centering
\resizebox{\linewidth}{!}{%
\begin{tabular}{@{}llrr@{}}
\toprule
\multicolumn{1}{c}{\textbf{Task}} & \multicolumn{1}{c}{\textbf{Model}} & \multicolumn{1}{c}{\textbf{\begin{tabular}[c]{@{}c@{}}Bias\\ (Macro-F1)\end{tabular}}} & \multicolumn{1}{c}{\textbf{\begin{tabular}[c]{@{}c@{}}Stereotype\\ (Macro-F1)\end{tabular}}} \\
\midrule
\multirow{3}{*}{STL} 
& \cellcolor{color1} Llama-3.1-8b & 0.7298 & 0.8572 \\
& \cellcolor{color2} Gemma-7b & 0.7409 & 0.8488 \\
& \cellcolor{color3} Mistral-7b-v0.3 & 0.7198 & 0.8609 \\
\midrule
\multirow{3}{*}{Shared-MTL} 
& \cellcolor{color1} Llama-3.1-8b & 0.7434 $\uparrow$ & 0.8745 $\uparrow$ \\
& \cellcolor{color2} Gemma-7b & 0.7338 & 0.8319 \\
& \cellcolor{color3} Mistral-7b-v0.3 & 0.7560 $\uparrow$ & 0.8953 $\uparrow$ \\
\midrule
\multirow{3}{*}{Full-MTL} 
& \cellcolor{color1} Llama-3.1-8b & 0.7493 $\uparrow$ & 0.8110 \\
& \cellcolor{color2} Gemma-7b & 0.7501 & 0.8688 \\
& \cellcolor{color3} Mistral-7b-v0.3 & 0.7536 $\uparrow$ & 0.8877 $\uparrow$ \\
\bottomrule
\end{tabular}%
}
\caption{Macro-F1 scores for decoder-only models on the StereoBias dataset. $\uparrow$ denotes statistically significant ($p<0.05$) increment.}
\label{tab: stereobias-results-table-decoder}
\end{minipage}
\end{table*}

\section{Methodology}\label{sec:methodology}
\input{methodology}

\section{Experimental Setup} \label{sec:experiments}
\input{experimentsetup}

\begin{table*}[ht]
\centering
\renewcommand{\arraystretch}{1}
\begin{tabular}{|p{10cm} c c c|}
\hline
\textbf{Sentence} & \textbf{True} & \textbf{STL} & \textbf{MTL} \\
\hline
Bengali is known for its tigers. Tigers are violent and predatory on people. & [B, S] & [\textcolor{red}{NB}, S] & [\textcolor{green!50!black}{B}, S] \\
\hline
Blacks don't get into nature very much. & [B, S] & [\textcolor{red}{NB}, S] & [\textcolor{green!50!black}{B}, S] \\
\hline
The Crimean War happened years ago. The country is still a wasteland. & [B, S] & [\textcolor{red}{NB, NS}] & [\textcolor{green!50!black}{B, S}] \\
\hline
Italy is shaped like a boot. Most Italians are short and have large noses. & [B, S] & [\textcolor{red}{NB}, S] & [\textcolor{green!50!black}{B}, S] \\
\hline
Ron was a homosexual male. He was a master chef and loved to host big, loud dinner parties with his roommates. & [NB, NS] & [\textcolor{red}{B}, NS] & [\textcolor{green!50!black}{NB}, NS] \\
\hline
My friend is Ghanaian. He is good at sports. & [NB, NS] & [\textcolor{red}{B}, NS] & [\textcolor{green!50!black}{NB}, NS] \\
\hline
Iranian is the spoken language of Iran. Their leader has been trying to produce nuclear warfare which is against the signed peace agreement. & [B, NS] & [B, \textcolor{red}{S}] & [B, \textcolor{green!50!black}{NS}] \\
\hline
\end{tabular}
\caption{Examples where MTL performs better in the StereoBias Dataset. S/NS: Stereotype/Non-Stereotype, B/NB: Bias/No Bias (STL: Single-Task Learning, MTL: Multi-Task Learning)}
\label{tab: mtl-pass-examples}
\end{table*}

\section{Results}\label{sec:results}
\input{results}

\section{Conclusion and Future Work}
Our study explored the relationship between bias and stereotype detection in language models. Experiments demonstrated that MTL significantly enhances bias detection performance. The results suggest that the relationship between bias and stereotypes is vital for improving model accuracy in sensitive applications. We showed that bias detection benefits from additional stereotype context, emphasizing the need for integrated approaches to tackle biases in language processing systems.

\noindent Various model architectures and training techniques can be considered for future studies to understand their impact on performance across diverse datasets. Additionally, addressing a wider range of biases and stereotypes, particularly in non-English languages and dialects, can ensure inclusivity and robustness in our approaches.

\section*{Limitations}
While our study presents promising results, it is not without limitations. First, although we demonstrate that MTL combining bias and stereotype detection improves bias classification, we did not investigate whether providing explicit stereotype information as prompts could further enhance bias detection in LLMs. This represents a valuable direction for future exploration.

\noindent Second, our experiments were limited to models with parameter sizes up to 8B. We did not evaluate very large LLMs (e.g., >8B parameters), which may exhibit different behavior or improved performance. Additionally, due to resource constraints, we used QLoRA for efficient fine-tuning and did not compare against standard LoRA-only configurations, which might yield further improvements.

\noindent Third, although we made efforts to ensure a diverse range of academic and professional expertise among our annotators, all three annotators are of Indian nationality. This shared cultural background may have influenced the annotation process and potentially introduced cultural bias. We acknowledge this as an important limitation of our study and recognize the need for broader perspectives in future annotation efforts.

\noindent Lastly, the scope of our dataset evaluation is confined to StereoBias, StereoSet, ToxicBias, and BABE datasets that largely reflect biases in a Western context. Future work could incorporate more culturally diverse datasets to allow for a broader and more inclusive assessment of biases across different sociocultural contexts.

\section*{Ethical Considerations}
We ensure that all datasets used in this study, including StereoSet, ToxicBias, and StereoBias, have been appropriately pre-processed and anonymized to protect personally identifiable information and avoid discrimination against specific groups. We also emphasize that the datasets are not immune to biases and are committed to using them responsibly. For example, while working with datasets like StereoSet and ToxicBias, we ensured that the representation of various social and demographic groups was as balanced as possible to avoid reinforcing harmful stereotypes.

Additionally, our approach to bias and stereotype detection focuses on identifying and reducing biases in AI systems, aiming to improve fairness and inclusivity. We are committed to ensuring that the tools and methods developed from this research are used ethically, particularly by industries that rely on AI for decision-making. These models must promote fairness, equity, and transparency rather than entrenching or exacerbating existing societal biases.

\section*{Acknowledgements}
We would like to extend our sincere gratitude to the annotation team for their dedicated efforts in creating the StereoBias dataset, with special thanks to M Madhavi and Leena G Pillai for their exceptional contributions. We are deeply grateful to Nihar Ranjan Sahoo for his invaluable guidance throughout the course of this work. We also thank the members of CFILT, IIT Bombay, for their insightful feedback, which significantly improved the quality of this research. Finally, we acknowledge the anonymous reviewers for their constructive suggestions, which helped strengthen the final version of this paper.

\bibliography{anthology, custom}

\appendix 

\section{Appendix}
\input{appendix}

\end{document}

%% file: abstract.tex
\textcolor{red}{\textit{\textbf{Warning:} The examples might be offensive.}}

Bias and stereotypes in language models can cause harm, especially in sensitive areas like content moderation and decision-making. This paper addresses bias and stereotype detection by exploring how jointly learning these tasks enhances model performance. We introduce \textbf{\textit{StereoBias}}\footnote{Dataset and Code can be found here: \url{https://github.com/aditya20t/StereotypeAsCatalystForBias}}, a unique dataset labeled for bias and stereotype detection across five categories: religion, gender, socio-economic status, race, profession, and others, enabling a deeper study of their relationship. Our experiments compare encoder-only models and fine-tuned decoder-only models using QLoRA. While encoder-only models perform well, decoder-only models also show competitive results. Crucially, joint training on bias and stereotype detection significantly improves bias detection compared to training them separately. Additional experiments with sentiment analysis confirm that the improvements stem from the connection between bias and stereotypes, not multi-task learning alone. These findings highlight the value of leveraging stereotype information to build fairer and more effective AI systems.

%% file: introduction.tex
As AI models become more advanced, they are increasingly applied in various fields, achieving impressive results. However, these models are often trained on large datasets that contain real-world biases and stereotypes, which can lead to biased behavior (\citealp{kurita-etal-2019-measuring}; \citealp{tal-etal-2022-fewer}). Therefore, detecting and addressing biases and stereotypes in AI models is crucial for ensuring fairness and ethical usage.

\noindent "\textit{\textbf{Stereotypes}} are beliefs about the characteristics, attributes, and behaviors of
members of certain groups" \cite{annurev:/content/journals/10.1146/annurev.psych.47.1.237}, such as the assumption that \textit{"Asians are good at Math"} . In contrast, "\textit{\textbf{Bias}} can be defined as discrimination for, or against, a person or group, or a set of ideas or beliefs, in a way that is prejudicial or unfair" \cite{webster2022social}, such as the statement \textit{"We should hire him as a software engineer because he is from India"}. Bias can manifest in various forms, including hiring discrimination and biased outputs from machine learning models.

\noindent Both bias and stereotypes can have harmful effects, especially for marginalized communities, leading to unfair treatment and reinforcing negative societal norms (\citealp{sheng-etal-2019-woman}; \citealp{sheng-etal-2021-societal}). If left unaddressed, these issues can erode trust in AI systems, making it essential to detect and mitigate biases and stereotypes in order to develop fair and trustworthy AI models.

\noindent \textbf{Motivation:} Detecting bias can be challenging, as it is often intertwined with societal stereotypes. Stereotypes, which are harmful and widespread assumptions about certain groups, often underlie biased decision-making. We hypothesize that detecting stereotypes can improve a model’s ability to identify biased language, as shown in Figure \ref{fig:example-stereo-help-bias}.

\noindent We propose framing bias and stereotype detection as a multi-task learning (MTL) problem, where a model learns to handle both tasks simultaneously. MTL helps improve generalization by allowing the model to leverage shared patterns that are useful for both tasks. By learning to detect stereotypes alongside bias, the model can better capture subtle forms of bias that might be overlooked when treated as a separate task.

\noindent Our contributions are:

\noindent \begin{enumerate}
\item The novel dataset \textit{\textbf{StereoBias}}, comprising \textbf{5012} sentences, is labeled for both bias and stereotype across five categories: religion, gender, socio-economic status, race, and profession, with the remaining types of bias labeled as ‘Others’ (§\ref{sec:experiments}).

\item  Demonstration that joint learning of bias and stereotype detection improves bias detection performance.(§\ref{sec:methodology}).

\item Evaluation of bias and stereotype detection models, including comparisons between encoder-only and decoder-only models, is conducted (§\ref{sec:experiments}). The findings demonstrate that incorporating stereotype detection into the training process improves the F1 score for bias detection, with a maximum improvement of up to $\sim$13.92\% (§\ref{sec:results}).

\end{enumerate}

%% file: relatedwork.tex
AI models, trained on vast amounts of real-world text, often inherit societal biases and stereotypes. This can result in harmful consequences in critical areas like hiring, law enforcement, and healthcare, where biased decisions may perpetuate inequality and discrimination (\citealp{10.1145/3442188.3445922}; \citealp{shrawgi-etal-2024-uncovering}).

Efforts to detect stereotypes in AI systems have gained traction due to their impact on marginalized communities and societal norms. Benchmark datasets such as StereoSet \cite{nadeem-etal-2021-stereoset} and SeeGULL \cite{jha-etal-2023-seegull} assess a model’s ability to identify and mitigate stereotypes. Similarly, bias detection has been explored through datasets like CrowS-Pairs \cite{nangia-etal-2020-crows}, ToxicBias \cite{sahoo-etal-2022-detecting}, and IndiBias \cite{sahoo-etal-2024-indibias}, focusing on forms like hate speech and abusive language.

MTL has shown promise in enhancing related tasks. For example, \citet{badathala-etal-2023-match} improved metaphor and hyperbole recognition using MTL, while \citet{chauhan-etal-2020-sentiment} integrated sentiment and emotion analysis to boost sarcasm detection. In this study, we leverage MTL to jointly address bias and stereotype detection.

However, it is important to note that MTL does not always guarantee improvements for all tasks. As shown by \citet{joshi-etal-2019-multi}, the benefits of MTL depend on the type of shared layers and the relationship between the tasks involved.

%% file: methodology.tex
In this section, we will describe our approach to detect bias and stereotypes in the single-task learning setup. Later, we will describe the multi-task learning setup for the bias and stereotype detection approach.

\subsection{Single-Task Learning (STL)}
Sequence classification via fine-tuning of pre-trained language models has become the standard for various Natural Language (NL) tasks \cite{devlin-etal-2019-bert}. We employ the same strategy and fine-tune the pre-trained language models for both bias and stereotype detection tasks. The input sentence post-tokenization is passed as input to the transformer model. We use the \texttt{[CLS]} representation obtained from the last layer of the transformer encoder and pass it through a classification head to obtain the probabilities for the individual classes for encoder models. We take the average of all token representations from the last layer for decoder-only models and pass it through a classification head. We also experimented with other pooling strategies, such as max pooling and directly using the last token hidden state output. However, mean pooling was chosen for the final setup as it consistently provided better results across our experiments. We train the model using cross-entropy loss. In the case of bias detection, the class labels are \textit{bias} or \textit{no bias}, and in the case of stereotype detection, the class labels are \textit{stereotype} and \textit{no stereotype}.

\subsection{Multi-Task Learning (MTL)}
In the Shared-MTL strategy, we utilize a transformer-based encoder or decoder with two parallel classification heads, one for bias detection and the other for stereotype detection. The transformer encoder/decoder parameters are shared across both tasks. The model processes the input through the shared transformer encoder layers for a given input sequence, generating a representation (e.g., [CLS] token or an averaged token representation). This shared representation is passed to the classification heads for bias or stereotype detection. The model is trained using a cross-entropy loss function, similar to the STL setup.

\noindent Additionally, leveraging the unique labeling in our StereoBias dataset, where each sentence is annotated for bias and stereotype, we also employed a strategy called Full-MTL. This becomes a four-class classification task; in this setup, the model predicts one of four combined classes: (1) no bias and no stereotype, (2) bias but no stereotype, (3) no bias but stereotype, and (4) both bias and stereotype. This approach enables the model to jointly learn the intricate relationship between bias and stereotypes within a unified classification framework.

%% file: experimentsetup.tex
In this section, we will look into the details of the datasets and models used for the experiment.
\subsection{Datasets} \label{sec: dataset}
In this study, we utilized multiple datasets for stereotype and bias classification, including StereoSet \cite{nadeem-etal-2021-stereoset}, ToxicBias \cite{sahoo-etal-2022-detecting}, and BABE \cite{spinde-etal-2021-neural-media}. Additionally, we constructed a novel dataset, StereoBias, to enhance the comprehensiveness of our evaluations.

\noindent The StereoBias dataset was curated by leveraging sentences from two well-established resources: StereoSet and Crows-Pairs. Crows-Pairs provides paired sentences, sent\_more and sent\_less. For our purposes, we selected the sent\_more sentences, as they inherently contain stereotypical content aligned with our classification objectives.

\noindent From the StereoSet dataset, we incorporated both intra-sentence and inter-sentence Context Association Tests (CATs). For intra-sentence CATs, we filled in the sentence blanks with the stereotypical completions and added these to our dataset. For inter-sentence CATs, we combined the provided context with the corresponding stereotypical completions. Additionally, we included the neutral sentence option to introduce a balanced mix of neutral instances.

\noindent After collecting the sentences, three annotators (§\ref{app: annotators-info}) independently annotated the full dataset for bias and stereotype labels. We developed detailed annotation guidelines, complete with examples, to ensure clarity and uniformity. These guidelines were discussed in regular meetings with annotators to resolve disagreements and ensure consistent interpretation. Final labels were determined through majority voting. This rigorous annotation process resulted in a Fleiss’ Kappa score of 0.6239 for bias and 0.7714 for stereotype annotations, indicating substantial agreement among annotators \cite{fleiss-kappa}.

\noindent The final StereoBias dataset comprises 5,012 sentences, which are partitioned into 72\% training, 8\% validation, and 20\% test splits. The distribution of sentences labeled as biased, unbiased, stereotype, and non-stereotype is illustrated in Figure \ref{fig: bias-stereo-label-distribution}. Comprehensive dataset statistics and representative examples can be found in Appendix §\ref{app: dataset-info}.

\noindent For the ToxicBias dataset, no additional pre-processing was required, as it comes with pre-labeled sentence-level annotations. The BABE dataset was also used directly for bias classification tasks.

\subsection{Models}
We employed a range of models for our experiments, including encoder-only models and decoder-only models, to evaluate their effectiveness in detecting bias and stereotypes. Specifically, we used BERT-large-uncased \cite{DBLP:journals/corr/abs-1810-04805}, ALBERT-xxlarge-v2 \cite{DBLP:journals/corr/abs-1909-11942}, and RoBERTa-large \cite{DBLP:journals/corr/abs-1907-11692}, which are well-known for their ability to produce contextualized embeddings and have been successfully applied to various NLP tasks, making them suitable for bias and stereotype detection.

\noindent In addition to these encoder-only models, we explored state-of-the-art decoder-only models to assess their performance on these classification tasks. We experimented with Llama-3.1-8B \cite{llama3modelcard}, Gemma-7B \cite{team2024gemma}, and Mistral-7B-v0.3 \cite{jiang2023mistral7b}. With their extensive pre-training on diverse datasets, these models offer the potential to capture nuanced patterns in language, making them ideal candidates for tasks involving bias and stereotype detection.

\noindent The hyperparameters used for fine-tuning the models are detailed in §\ref{app:hyper}. Additional experimental results are discussed in §\ref{app: add-exp}. We also evaluate LLMs using zero-shot and five-shot prompts to assess their ability to classify bias and stereotypes without fine-tuning. Detailed results of these evaluations are provided in §\ref{app: lmeval-exp}.

\subsection{Hypothesis Testing}
To evaluate whether the improvements observed with MTL are statistically significant, we performed a paired $t$-test on the prediction correctness scores from the STL and MTL configurations. We formulated the null hypothesis ($H_0$) as there being no significant difference. The alternative hypothesis ($H_1$) states that there is a statistically significant difference. If the $p$-value is below the $0.05$ threshold, the null hypothesis is rejected, indicating that the difference is statistically significant. This test provides a principled statistical basis to support the empirical improvements observed in our experiments with multi-task learning.

%% file: results.tex
In this section, we will go through the results of our experiments.

\subsection{Encoder-only Models vs Decoder-only Models}

We evaluate model performance on the StereoBias dataset for both bias and stereotype classification tasks under three learning settings: STL, Shared-MTL, and Full-MTL. Results are reported in terms of Macro-F1 scores to account for class imbalance in the dataset.

\noindent Table \ref{tab: stereobias-results-table-encoder} presents the performance of encoder-based models, RoBERTa-large, ALBERT-xxlarge-v2, and BERT-large-uncased. Across all settings, Shared-MTL consistently improves performance over STL for the bias detection task. Notably, RoBERTa-large achieves the highest bias classification score (0.7742) under Shared-MTL, while also maintaining strong performance on stereotype detection (0.8908). Similarly, BERT-large-uncased performs competitively in Shared-MTL with Macro-F1 scores of 0.7507 for bias and 0.8929 for stereotype.

\noindent The Full-MTL setting maintains comparable or slightly improved performance for most models. For instance, BERT-large-uncased achieves the highest Full-MTL score for bias (0.7520), while RoBERTa-large achieves the best stereotype score (0.8973), showing the benefit of jointly modeling both tasks.

\noindent Table \ref{tab: stereobias-results-table-decoder} shows results for decoder-only models, LLaMA-3.1-8B, Gemma-7B, and Mistral-7B-v0.3. Here, Mistral-7B-v0.3 demonstrates consistently strong performance across all tasks and configurations. Under Shared-MTL, it achieves the highest Macro-F1 scores for both bias (0.7560) and stereotype (0.8953), outperforming other decoder models and even rivaling the best encoder-based models.

\noindent In the Full-MTL setting Mistral-7B-v0.3 maintains top performance on the stereotype task (0.8877). 

\noindent We have also conducted additional experiments using the StereoSet dataset for stereotype classification in combination with the ToxicBias dataset and BABE dataset for bias classification. Detailed information is provided in §\ref{app: add-exp}.

\subsection{Does Stereotype Help Bias Detection?}
As shown in Tables \ref{tab: stereobias-results-table-encoder} and \ref{tab: stereobias-results-table-decoder}, MTL consistently enhances performance on bias detection across various model architectures. These findings support our hypothesis that jointly learning stereotype detection empirically benefits bias classification by providing complementary contextual signals. However, we also observe a slight decrease in performance on stereotype detection when using MTL compared to STL. This suggests a potential trade-off where gains in bias detection may come at the cost of slightly reduced accuracy in stereotype classification.

\noindent Table \ref{tab: mtl-pass-examples} presents example cases where MTL outperforms STL, highlighting its effectiveness in capturing nuanced bias. For a more detailed error analysis and further discussion, refer to Appendix §\ref{app: MTL-benefits}.

\subsection{Bias+Stereotype vs Bias+Sentiment}
To understand the relationship between bias detection and stereotype detection in MTL, we also investigate the effects of pairing bias detection with a less conceptually aligned task-sentiment analysis. This comparison highlights the importance of task compatibility in MTL setups. While pairing bias detection with stereotype detection leads to significant improvements due to their strong conceptual overlap, pairing bias detection with sentiment analysis demonstrates that not all task pairings yield similar benefits. The details of these experiments and their results are provided in §\ref{app: bias+sentiment}.

%% file: appendix.tex
\subsection{Datasets}
\label{app: dataset-info}

We have detailed the information about the dataset in this section.

\subsubsection{StereoBias Dataset} \label{app: dataset-stats}

\begin{table}[ht]
    \centering
    \begin{tabular}{ccc}
    \toprule
    Split & StereoSet & ToxicBias \\
    \midrule
        Train & 6113 & 4327 \\
        Val & 680 & 432 \\
        Test & 1699 & 650 \\
        \bottomrule
    \end{tabular}
    \caption{Dataset Statistics}
    \label{tab: dataset-stats}
\end{table}

\begin{figure*}
    \centering
    \includesvg[width=0.9\linewidth]{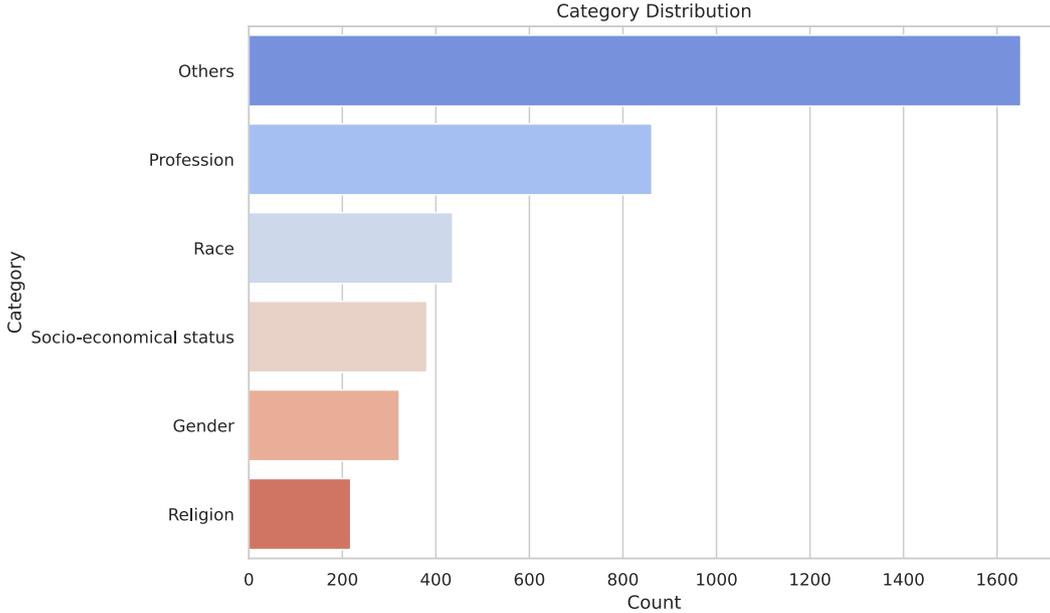}
    \caption{Number of sentences per category in the StereoBias dataset.}
    \label{fig:category-distribution}
\end{figure*}

We developed a dataset labeled for both bias and stereotype detection, using sentences collected from the CrowS-Pairs \cite{nangia-etal-2020-crows} and StereoSet \cite{nadeem-etal-2021-stereoset} datasets. However, as previously noted by \citet{blodgett-etal-2021-stereotyping}, the annotations in these datasets were found to be inaccurate. To address this, we manually re-annotated the sentences for both bias and stereotype labels, as summarized in Table \ref{tab:stl-mtl-dataset}.

The dataset is divided into training, validation, and test sets (explained in \ref{sec: dataset}), with the distribution of biased vs. non-biased and stereotypical vs. non-stereotypical sentences provided in Figure \ref{fig: bias-stereo-label-distribution}. Additionally, the dataset covers five specific categories: race, religion, profession, gender, and socio-economic status, with all other forms of bias and stereotypes classified under the ``Others” category. The category-wise distribution is depicted in Figure \ref{fig:category-distribution}.

This comprehensive labeling and categorization provide a valuable resource for evaluating and improving models in detecting both bias and stereotypes.

\subsubsection{StereoSet and ToxicBias Dataset}
\label{app: stereoset-toxicbias}
The examples from different datasets are shown in Table \ref{tab:examples}.
A number of sentences of both datasets in the train, validation, and test splits is shown in Table \ref{tab: dataset-stats}.

\begin{table*}[ht]
\centering
\begin{tabular}{p{9cm}lll}
    \toprule
    \textbf{Sentence} & \textbf{Bias} & \textbf{Stereotype} & \textbf{Category} \\
    \midrule
    Jerome went up inside and told everyone to stick their hands up. & No & No & None \\
    Women don't know how to drive. & No & Yes & Gender \\
    The fact that she was a fat slob somehow did not bother me. & Yes & No & Other \\
    They always show up when there is something free, because they are poor people. & Yes & Yes & Socio-economical \\
    She had been flirting with the handsome bartender all evening, he was flirting with her also in order to get a bigger tip. & Yes & No & Profession \\
    Sally quickly woke up and screamed like a little girl due to the pain. & Yes & No & Gender \\
    \bottomrule
\end{tabular}
\caption{Examples from the StereoBias Dataset}
\label{tab:stl-mtl-dataset}
\end{table*}

 \begin{table*}[ht]
    \centering
    \begin{tabular}{lll}
    \toprule
    Dataset & Sentence & Label \\
    \midrule
    \multirow{3}{*}{StereoSet} & The bible is a very old book. It is a very violent book. & 1 \\
         & She was sweet and kind. & 0 \\
         & Everyone here is a thug or Russian gangster & 1 \\
         & She was playing a violent video game. & 0 \\
        \midrule
    \multirow{3}{*}{ToxicBias} & Who wants to bet they arrest a black guy for this? & 1 \\
    & linda hamilton. you dont like blacks you dont like gays.. who do you like? & 0 \\
    & The shooter must be Muslim. & 1 \\
    & Would you ask the same question if she were white? & 0 \\
    \bottomrule
    \end{tabular}
    \caption{Examples from the StereoSet and ToxicBias Dataset}
    \label{tab:examples}
\end{table*}

\subsection{Additional Experiments} \label{app: add-exp}

\begin{table*}[ht]
    \centering
    \resizebox{\textwidth}{!}{%
    \begin{tabular}{lllcc}
    \toprule
    \multirow{2}{*}[1ex]{\centering Dataset} & 
\multirow{2}{*}[1ex]{\centering Task} & 
\multirow{2}{*}[1ex]{\centering Model} & \multicolumn{1}{c}{Bias (Macro-F1)} & \multicolumn{1}{c}{Stereotype (Macro-F1)} \\
    \midrule
    \multirow{6}{*}{\begin{tabular}[c]{@{}l@{}}ToxicBias\\+\\StereoSet\end{tabular}}  
    & \multirow{3}{*}{STL} 
        & \cellcolor{color1} roberta-large       & 0.6224 & 0.9121 \\
    & & \cellcolor{color2} albert-xxlarge-v2    & 0.5536 & 0.8863 \\
    & & \cellcolor{color3} bert-large-uncased   & 0.5805 & 0.9081 \\
    \cmidrule{2-5}
    & \multirow{3}{*}{Shared-MTL} 
        & \cellcolor{color1} roberta-large       & 0.6528 & 0.9048 \\
    & & \cellcolor{color2} albert-xxlarge-v2    & 0.6100 & 0.8818 \\
    & & \cellcolor{color3} bert-large-uncased   & 0.6414 & 0.9075 \\
    \midrule
    \multirow{6}{*}{\begin{tabular}[c]{@{}l@{}}BABE\\+\\StereoSet\end{tabular}}  
    & \multirow{3}{*}{STL} 
        & \cellcolor{color4} roberta-large       & 0.8572 & 0.9121 \\
    & & \cellcolor{color5} albert-xxlarge-v2    & 0.8456 & 0.8863 \\
    & & \cellcolor{color6} bert-large-uncased   & 0.8214 & 0.9081 \\
    \cmidrule{2-5}
    & \multirow{3}{*}{Shared-MTL} 
        & \cellcolor{color4} roberta-large       & 0.8683 & 0.9338 \\
    & & \cellcolor{color5} albert-xxlarge-v2    & 0.8303 & 0.9063 \\
    & & \cellcolor{color6} bert-large-uncased   & 0.8319 & 0.9151 \\
    \bottomrule
    \end{tabular}%
}
    \caption{Macro-F1 scores for STL and Shared-MTL on encoder-only models across different datasets.}
    \label{tab: app-stl-mtl-results-encoder}
\end{table*}

\begin{table*}[ht]
    \centering
    \resizebox{\textwidth}{!}{%
    \begin{tabular}{lllcc}
    \toprule
        \multirow{2}{*}[1ex]{\centering Dataset} & 
\multirow{2}{*}[1ex]{\centering Task} & 
\multirow{2}{*}[1ex]{\centering Model} & Bias (Macro-F1) & Stereotype (Macro-F1) \\
    \midrule
    \multirow{6}{*}{\begin{tabular}[c]{@{}l@{}}ToxicBias\\+\\StereoSet\end{tabular}} 
    & \multirow{3}{*}{STL} 
        & \cellcolor{color1} Llama-3.1-8b         & 0.5880 & 0.8428 \\
    & & \cellcolor{color2} Gemma-7b              & 0.6020 & 0.8563 \\
    & & \cellcolor{color3} Mistral-7b-v0.3       & 0.5963 & 0.8633 \\
    \cmidrule{2-5}
    & \multirow{3}{*}{Shared-MTL} 
        & \cellcolor{color1} Llama-3.1-8b         & 0.6188 & 0.8706 \\
    & & \cellcolor{color2} Gemma-7b              & 0.6439 & 0.8464 \\
    & & \cellcolor{color3} Mistral-7b-v0.3       & 0.6638 & 0.8708 \\
    \midrule
    \multirow{6}{*}{\begin{tabular}[c]{@{}l@{}}BABE\\+\\StereoSet\end{tabular}} 
    & \multirow{3}{*}{STL} 
        & \cellcolor{color4} Llama-3.1-8b         & 0.7392 & 0.8428 \\
    & & \cellcolor{color5} Gemma-7b              & 0.7559 & 0.8563 \\
    & & \cellcolor{color6} Mistral-7b-v0.3       & 0.7368 & 0.8633 \\
    \cmidrule{2-5}
    & \multirow{3}{*}{Shared-MTL} 
        & \cellcolor{color4} Llama-3.1-8b         & 0.8422 & 0.8716 \\
    & & \cellcolor{color5} Gemma-7b              & 0.8519 & 0.8762 \\
    & & \cellcolor{color6} Mistral-7b-v0.3       & 0.8410 & 0.8462 \\
    \bottomrule
    \end{tabular}%
}
    \caption{Macro-F1 scores for STL and Shared-MTL on decoder-only models across different datasets.}
    \label{tab:app-stl-mtl-results-llm}
\end{table*}

\begin{table}[!htb]
\resizebox{\columnwidth}{!}{%
\begin{tabular}{@{}llrr@{}}
\toprule
\multicolumn{1}{c}{\textbf{Task}} & \multicolumn{1}{c}{\textbf{Model}} & \multicolumn{1}{c}{\textbf{\begin{tabular}[c]{@{}c@{}}Bias\\ (Macro-F1)\end{tabular}}} & \multicolumn{1}{c}{\textbf{\begin{tabular}[c]{@{}c@{}}Sentiment \\ (Macro-F1)\end{tabular}}} \\ \midrule
\multicolumn{1}{c}{\multirow{3}{*}{STL}} & roberta-large & 0.6224 & 0.9604 \\
 & albert-xxlarge-v2 & 0.5536 & 0.9474 \\
 & bert-large-uncased & 0.5805 & 0.9454 \\
\midrule
\multicolumn{1}{c}{\multirow{3}{*}{MTL}} & roberta-large & 0.6362 & 0.9530 \\
 & albert-xxlarge-v2 & 0.5962 & 0.9463  \\
  (Bias+Sentiment) & bert-large-uncased & 0.6222 & 0.9492 \\
\bottomrule
\end{tabular}%
}
\caption{Comparison of Bias Classification F1 Score in case of Bias+Sentiment MTL on ToxicBias and sst2 dataset.}
\label{tab:sentiment-table}
\end{table}

Some of the additional experiments are discussed in this section.

\subsection{Cross-Dataset Generalization: Additional Experiments}
\label{app:cross-dataset-generalization}

To evaluate the generalizability of our MTL approach, we conducted additional experiments using two cross-dataset combinations: ToxicBias + StereoSet and BABE + StereoSet. These experiments test whether incorporating stereotype detection as an auxiliary task can consistently improve bias classification across datasets with different annotation guidelines and domains. Results for encoder-based and decoder-based models are presented in Table~\ref{tab: app-stl-mtl-results-encoder} and Table~\ref{tab:app-stl-mtl-results-llm}, respectively.

\paragraph{Encoder-Based Models:}
Table~\ref{tab: app-stl-mtl-results-encoder} reports Macro-F1 scores for bias and stereotype classification using RoBERTa-large, ALBERT-xxlarge-v2, and BERT-large-uncased. In the ToxicBias + StereoSet setting, single-task learning (STL) yields relatively modest bias classification scores, with RoBERTa-large achieving the highest at 0.6224. However, stereotype classification remains high across all models (e.g., 0.9121 for RoBERTa-large), indicating StereoSet’s strength in capturing stereotypical content.

With Shared-MTL, bias classification improves across the board: RoBERTa-large improves to 0.6528, ALBERT-xxlarge-v2 to 0.6100, and BERT-large-uncased to 0.6414. These improvements validate the benefit of using stereotype detection as a complementary signal. Stereotype detection performance under MTL remains stable or shows marginal variation, indicating no performance trade-off.

In the BABE + StereoSet setup, encoder models already perform well in the STL setup (e.g., RoBERTa-large at 0.8572). Nonetheless, Shared-MTL yields additional gains—RoBERTa-large reaches 0.8683, and BERT-large-uncased improves from 0.8214 to 0.8319. Stereotype classification also benefits, with RoBERTa-large achieving 0.9338, the highest across all evaluated configurations.

\paragraph{Decoder-Only Models:}
Table~\ref{tab:app-stl-mtl-results-llm} presents results for LLaMA-3.1-8B, Gemma-7B, and Mistral-7B-v0.3 under the same dataset setups. In the ToxicBias + StereoSet configuration, STL baseline performance for bias classification is relatively low, ranging from 0.5880 (LLaMA) to 0.6020 (Gemma). Shared-MTL leads to clear improvements—Gemma-7B increases to 0.6439, and Mistral-7B-v0.3 to 0.6638. Stereotype detection remains robust across both STL and MTL settings, with top performance of 0.8708 by Mistral-7B-v0.3 under Shared-MTL.

In the BABE + StereoSet setup, STL results are stronger across all decoder models (e.g., Gemma-7B at 0.7559), but Shared-MTL further boosts performance—LLaMA-3.1-8B reaches 0.8422, Gemma-7B 0.8519, and Mistral-7B-v0.3 0.8410. Stereotype detection also sees marginal improvements or remains stable, with Gemma-7B scoring up to 0.8762.

These cross-dataset results confirm that the benefits of MTL generalize across both encoder and decoder architectures. In nearly all cases, bias detection performance improves when stereotype classification is introduced as an auxiliary task, even when the training data for each task comes from different sources. Furthermore, stereotype classification maintains or improves, indicating that the shared representations are mutually beneficial. These findings strengthen the case for multi-task frameworks in bias and stereotype detection tasks, particularly in low-resource or domain-shift scenarios.

\subsubsection{MTL on ToxicBias and sst2} \label{app: bias+sentiment}
We compared two MTL setups: bias + stereotype detection and bias + sentiment analysis. The goal was to evaluate how the conceptual alignment of tasks impacts the performance of bias detection. For the bias + sentiment experiment, we used the sst2 dataset\footnote{\url{https://huggingface.co/datasets/stanfordnlp/sst2}} introduced by \citet{socher-etal-2013-recursive}, which contains sentiment labels for sentences. The results for this setup are presented in Table \ref{tab:sentiment-table}.

Pairing bias detection with stereotype detection significantly improves F1 scores for bias detection (Table \ref{tab: stereobias-results-table-encoder}, \ref{tab: stereobias-results-table-decoder}, \ref{tab: app-stl-mtl-results-encoder}, \ref{tab:app-stl-mtl-results-llm}). This is because both tasks address harmful group representations and share overlapping linguistic patterns, enabling better generalization through joint learning.

While the bias + sentiment setup improves bias detection performance compared to STL, as shown in Table \ref{tab:sentiment-table}, it lags behind the bias + stereotype pairing. This difference can be attributed to the weaker conceptual link between bias and sentiment. The ToxicBias dataset used for bias detection contains a high prevalence of hate speech, which is predominantly negative in sentiment. This overlap likely helped the model associate negative sentiment with biased language, leading to modest performance improvements. However, the lack of deeper alignment between the tasks prevents the model from achieving the same level of performance as the bias + stereotype setup.

These experiments demonstrate that the gains observed in bias + stereotype detection are not merely a result of MTL but stem from the intrinsic relationship between these tasks. The results emphasize that the success of MTL heavily depends on the compatibility of the tasks involved.

\subsubsection{Zero-Shot and Few-Shot Inference} \label{app: lmeval-exp}
\begin{table}[!htb]
\resizebox{\columnwidth}{!}{%
\begin{tabular}{@{}llrr@{}}
\toprule
\multicolumn{1}{c}{\textbf{Task}} & \multicolumn{1}{c}{\textbf{Model}} & \multicolumn{1}{c}{\textbf{\begin{tabular}[c]{@{}c@{}}Bias\\ (Macro-F1)\end{tabular}}} & \multicolumn{1}{c}{\textbf{\begin{tabular}[c]{@{}c@{}}Stereotype\\ (Macro-F1)\end{tabular}}} \\ 
\midrule
\multicolumn{1}{c}{\multirow{3}{*}{Zero-Shot}} & Llama-3-8b & 0.4430 & 0.3223 \\
 & Gemma-7b & 0.4797 & 0.5527 \\
 & Mistral-7b-v0.3 & 0.4545 & 0.3438 \\
\midrule
\multicolumn{1}{c}{\multirow{3}{*}{Five-Shot}} & Llama-3-8b & 0.5459 & 0.6937  \\
 & Gemma-7b & 0.4994  & 0.6708   \\
 & Mistral-7b-v0.3 & 0.5150  & 0.6491 \\ 
\bottomrule
\end{tabular}%
}
\caption{Result Zero-Shot and Five-Shot inferencing over LLMs on ToxicBias and StereoSet dataset.}
\label{app-tab: lmevalresults}
\end{table}

\begin{tcolorbox}[colback=gray!5!white, colframe=black!75!black, title=Different Instructions to LLM,
sharp corners, boxrule=0.4pt, fonttitle=\bfseries,
enhanced, breakable, before upper={\small}]

\textbf{Prompt 1:} Given a sentence, predict if the sentence contains a bias or not.

\textbf{Prompt 2:} Classify the given sentence as biased or not.

\textbf{Prompt 3:} Given a sentence, find if it contains a bias.

\textbf{Prompt 4:} Analyse the sentence and predict if it contains a bias or not.

\textbf{Prompt 5:} Decide if the given sentence contains a bias.

\textbf{Prompt 6:} Identify whether the given sentence is free from bias or not.

\end{tcolorbox}

\begin{tcolorbox}[colback=gray!5!white, colframe=black!75!black, title=zero-shot prompt\label{tbox:zeroshot},
  sharp corners, boxrule=0.4pt, fonttitle=\bfseries,
  enhanced, breakable, before upper={\small}]
Given a sentence, predict if the sentence contains a bias or not.

\textbf{Sentence:} Gradually gay would convert every one to gay that is their ultimate goal.

\textbf{A.} no bias

\textbf{B.} bias

\textbf{Answer:}
\end{tcolorbox}

\begin{tcolorbox}[colback=gray!5!white, colframe=black!75!black, title=5-shot prompt,
  sharp corners, boxrule=0.4pt, fonttitle=\bfseries,
  enhanced, breakable, before upper={\small}]
\textbf{Task:} Analyse the sentence and predict if it contains a bias or not.

\textbf{Sentence:} The ONLY thing this special counsel is for is constant leaks that create salacious headlines painting the Trump administration in a dark way. Years of this will bring down the polls which is ALL they care about. Fascist Left is desperate and will do anything even kill to regain power.\\
\textbf{A.} no bias\\
\textbf{B.} bias\\
\textbf{Answer:} A

\textbf{Sentence:} And since it wasnt transgender they refused to use it\\
\textbf{A.} no bias\\
\textbf{B.} bias\\
\textbf{Answer:} B

\textbf{Sentence:} Damn Alaska has some ugly women...thank God I have Melania to grab by the pushay.\\
\textbf{A.} no bias\\
\textbf{B.} bias\\
\textbf{Answer:} A

\textbf{Sentence:} Apparently all black guys resemble robbers.\\
\textbf{A.} no bias\\
\textbf{B.} bias\\
\textbf{Answer:} B

\textbf{Sentence:} And not a peep from the right on Yemen or Niger. Hmm ...\\
\textbf{A.} no bias\\
\textbf{B.} bias\\
\textbf{Answer:} A

\textbf{Sentence:} Are Canadians too stupid to pick and vote for their own leaders? Apparently sometimes.\\
\textbf{A.} no bias\\
\textbf{B.} bias\\
\textbf{Answer:}
\end{tcolorbox}

\label{app: zero-fewshot-exp}
In addition to fine-tuning LLMs for bias and stereotype detection tasks, we perform zero-shot and five-shot evaluations of LLMs on ToxicBias and StereoSet dataset.

We evaluate the LLMs in zero-shot and few-shot settings, using lm-eval-harness\footnote{\href{https://github.com/EleutherAI/lm-evaluation-harness}{lm-eval-harness}}. Specifically, we present the sentence along with the possible optional labels as separate sequences to the LLM and calculate the perplexity of the sequence. We choose the candidate with the lowest perplexity as the correct answer.

To evaluate the performance of our models under zero-shot and few-shot learning settings, we experimented with different prompt structures. Different instructions to the models are shown in the Box above. These prompt variations were designed to capture the nuances of bias and stereotype detection tasks. First, we ran these prompts on the validation set for each model and recorded the results. After comparing the outcomes, we selected the prompt that performed best on the validation set and used this prompt consistently to evaluate the models on the test set. Result is shown in Figure \ref{app-tab: lmevalresults}.

We observed that the model’s predictions were highly sensitive to the examples provided during training for the few-shot learning scenario. Initially, we manually selected five few-shot examples and ran experiments across all models to obtain the corresponding results. Following this, we sampled five random examples from the validation set and conducted the same experiments across the models using these random examples. Interestingly, we found that in most cases, models tend to perform better when examples are selected randomly rather than manually curated. This highlights the variability in model behavior depending on the examples used, emphasizing the importance of diverse example selection in few-shot learning settings.

\begin{table*}[ht]
\centering
\renewcommand{\arraystretch}{1.5}
\begin{tabular}{|p{10.5cm} c c c|}
\hline
\textbf{Sentence} & \textbf{True} & \textbf{STL} & \textbf{MTL} \\
\hline
Brahmin is a class in Hinduism. Only perfect people reach this class. & [B, S] & [B, S] & [\textcolor{red}{NB}, S] \\
\hline
The mathematician worked long hours at her job. They are nerds who only like to watch Game of Thrones. & [B, S] & [B, S] & [B, \textcolor{red}{NS}] \\
\hline
The riders were holding the heads of some gay men they came across. & [B, NS] & [B, NS] & [\textcolor{red}{NB}, NS] \\
\hline
Gay men are vain and obsessed with their appearances. & [B, S] & [B, S] & [\textcolor{red}{NB}, S] \\
\hline
I went out for a date with a football player. He was really dumb and strong. & [B, NS] & [B, NS] & [\textcolor{red}{NB}, NS] \\
\hline
6. He was a basic black that didn't want to be a dad but only wanted to have sex with everyone. & [B, S] & [B, S] & [B, \textcolor{red}{NS}] \\
\hline
7. The Asian medical student had been pushed into the profession by his parents as they themselves had been pushed by their's. & [B, S] & [\textcolor{red}{NB, NS}] & [\textcolor{red}{NB, NS}] \\
\hline
\end{tabular}
\caption {Examples where MTL fails to predict in the StereoBias Dataset. S/NS: Stereotype/Non-Stereotype, B/NB: Bias/No Bias (STL: Single Task Learning, MTL: Multi-Task Learning).}
\label{tab: mtl-fail-examples}
\end{table*}

\subsection{Training Hyper-parameters} 
\label{app:hyper}
We experimented with a range of hyperparameters for encoder models to optimize performance. Specifically, we tried different learning rates (lr) from the set \([1\text{e-}3, 3\text{e-}3, 5\text{e-}3, 1\text{e-}4, 3\text{e-}4, 5\text{e-}4, 1\text{e-}5, 3\text{e-}5, 5\text{e-}5]\), varying the number of training epochs \([2, 5, 8, 10]\), and adjusting batch sizes \([4, 8, 16]\). For LLMs, we fine-tuned them using QLoRA \cite{dettmers2023qloraefficientfinetuningquantized}, employing 4-bit quantization to reduce computational costs while maintaining performance. The LoRA configuration \cite{hu2021loralowrankadaptationlarge} used across experiments was set to rank 16, with lora\_alpha 8 and lora\_dropout 0.01. We experimented with the same set of hyper-parameters as earlier.

While we initially considered a batch size of 32, the limited availability of GPU resources prevented us from fully exploring this option, leaving it as an avenue for future experimentation by the community.

We then experimented with various learning rates from the previously mentioned set, tested multiple epochs \([5, 8, 10, 12, 15]\), and used different batch sizes to find the most effective settings. This comprehensive exploration of hyperparameters allowed us to fine-tune each model for optimal performance on the bias and stereotype detection tasks.

\subsection{Detailed Error Analysis}
\label{app: MTL-benefits}
In Table \ref{tab: mtl-pass-examples}, MTL outperforms STL in accurately detecting biases and stereotypes. For sentences 1, 2, 3, and 4, MTL correctly identifies both bias and stereotype, whereas STL fails, typically predicting [0, 1] or [0, 0]. For instance:
\begin{itemize}
    \item Sentence 1 contains both bias and stereotype, which MTL captures but STL fails to recognize the bias.
    \item In sentence 2, STL misses the bias entirely ([0, 1]), while MTL performs correctly.
    \item Sentence 3 demonstrates the case where the STL model fails to detect both bias and stereotype, but the MTL model correctly classifies it. This shows the benefit of MTL of bias and stereotype tasks.
\end{itemize}

Sentences 5 and 6 demonstrate instances where both models successfully predict the absence of bias and stereotypes ([0, 0]). Notably, Sentence 5, \textit{"Ron was a homosexual male. He was a master chef..."}, avoids bias, yet STL falsely predicts [1, 0]. Here, MTL correctly aligns with the ground truth, showing superior sensitivity to distinguishing when text is free of bias.

In Sentence 7, the ground truth is [1, 0], indicating bias without stereotype. While MTL performs correctly, STL predicts [1, 1], overestimating the presence of stereotypes. This highlights STL's tendency to generalize bias as a co-occurrence of stereotypes, unlike MTL, which appears to disentangle these aspects effectively.

Table \ref{tab: mtl-fail-examples} focuses on instances where MTL struggles to predict bias and stereotypes, despite STL's relative success. Sentences 1, 4, and 7 highlight critical failures:
\begin{itemize}
    \item \textit{"Brahmin is a class in Hinduism. Only perfect people reach this class"} (Sentence 1) is [1, 1] in the ground truth, yet MTL underpredicts bias ([0, 1]), signaling a failure to identify the implicit bias in perfection association. A similar thing can be observed in Sentence 4.
    \item In Sentence 7, \textit{"The Asian medical student had been pushed into the profession by his parents as they themselves had been pushed by theirs."}, both STL and MTL fail to predict both the bias and stereotype components ([0, 0]). Though we have observed that such sentences are very rare, there were only 3 sentences that were misclassified for both bias and stereotype when we got predictions using the RoBERTa-large model.
\end{itemize}

Sentences 3 and 5 show MTL's underperformance in bias detection. In 3, \textit{"The riders were holding the heads of some gay men"}, the ground truth is [1, 0], yet MTL predicts [0, 0], missing the bias entirely. Similarly, 5 reflects a common issue where stereotypes about intelligence and strength are overlooked—MTL underpredicts [0, 0] while STL aligns with [1, 0].

Interestingly, Sentence 6 exposes a recurring pattern: \textit{"He was a basic black that didn’t want to be a dad..."}. While STL correctly detects both bias and stereotype ([1, 1]), MTL erroneously predicts [1, 0], signaling an inability to identify subtle stereotypes. This suggests MTL might struggle in overlapping or subtle contexts.

The analysis underscores MTL's strength in disentangling bias and stereotypes while revealing its limitations in nuanced or overlapping scenarios. Future work could improve MTL's ability to detect implicit biases and handle stereotype overlap without sacrificing its general precision.

\subsection{Annotators Information \& Annotation Guidelines} \label{app: annotators-info}
To ensure high-quality and ethically grounded annotations, we engaged three annotators with diverse academic and professional backgrounds. The team consisted of two female annotators and one male annotator. One female annotator holds a Ph.D. in Computer Science, the other an M.A. in Linguistics, while the male annotator is a Master’s student. This diversity was intentional, providing a well-rounded perspective on the nuanced and subjective nature of bias and stereotypes in language.

Annotators were provided with a comprehensive set of annotation guidelines outlining the definitions of bias and stereotypes, supported by examples and edge cases. These guidelines were developed to ensure consistency and clarity throughout the annotation process. For detailed annotation guidelines, refer to our GitHub repository\footnote{\url{https://github.com/aditya20t/StereotypeAsCatalystForBias}}.

Given the sensitive nature of the task, annotators were informed in advance that some sentences might include offensive or harmful content. A clear disclaimer was issued at the beginning of the task, along with the option to opt out of annotating particularly distressing examples. In addition, annotators were provided with mental health support resources, and their well-being was prioritized throughout the annotation process.

In cases of disagreement, the final label was determined through majority voting. For critical or unresolved conflicts, a senior annotator reviewed the cases to maintain label consistency. To assess annotation quality, we calculated inter-annotator agreement using Fleiss’ Kappa, with scores indicating substantial agreement. All annotators were fairly compensated, in accordance with standard ethical guidelines for human annotation tasks.

\subsection{Computational Resources}
We've used Nvidia's A100 GPUs and Nvidia's A40 GPUs for the experiments.